\documentclass{article}%{sn-jnl}% APA Reference Style 

\usepackage{arxiv}
\usepackage{comment}
\usepackage{graphicx}%
\usepackage{multirow}%
\usepackage{amsmath,amssymb,amsfonts}%
\usepackage{amsthm}%
\usepackage{mathrsfs}%
\usepackage[title]{appendix}%
\usepackage{xcolor}%
\usepackage{manyfoot}%
\usepackage{booktabs}%
\usepackage{algorithm}%
\usepackage{algorithmicx}%
\usepackage{algpseudocode}%
\usepackage{listings}%
\usepackage{soul}
\usepackage{natbib}
\usepackage{arabtex}
\usepackage[T1]{fontenc}
\usepackage[utf8]{inputenc}
\usepackage{tipa}
\usepackage{array}
\usepackage{hyperref}
\usepackage{utf8}
\setcode{utf8}

\title{Building Foundations for Natural Language Processing of Historical Turkish: Resources and Models}

\author{\textbf{\c{S}aziye Bet\"{u}l \"{O}zate\c{s}}*\\ 
Boğaziçi University\\
\small{\texttt{saziye.ozates@bogazici.edu.tr}}
\And
Tar{\i}k Emre T{\i}ra\c{s}$^{\ddagger}$\\
Boğaziçi University\\
\small{\texttt{tarik.tiras@std.bogazici.edu.tr}}
\And
Ece Elif Adak$^{\ddagger}$\\
Boğaziçi University\\
\small{\texttt{ece.adak@std.bogazici.edu.tr}}
\And
Berat Do\u{g}an$^{\ddagger}$\\
Boğaziçi University\\
\small{\texttt{berat.dogan@std.bogazici.edu.tr}}
\And
Fatih Burak Karag\"{o}z
\\
Boğaziçi University\\
\small{\texttt{fatih.karagoz@std.bogazici.edu.tr}}
\And
Efe Eren Gen\c{c}\\
Saarland University\\
\small{\texttt{efge00001@stud.uni-saarland.de}}
\And
Esma F. Bilgin Ta\c{s}demir
\\
Medeniyet University\\
\small{\texttt{esmabilgin.tasdemir@medeniyet.edu.tr}}
}
\date{}

\usepackage{etoolbox}
\pretocmd\document{\endgroup}{}{\fail} % close document group
\begin{document}
\maketitle

\abstract{
This paper introduces foundational resources and models for natural language processing (NLP) of historical Turkish, a domain that has remained underexplored in computational linguistics. We present the first named entity recognition (NER) dataset, {\it HisTR}, and the first Universal Dependencies treebank, {\it OTA-BOUN}, for a historical form of the Turkish language along with transformer-based models trained using these datasets for named entity recognition, dependency parsing, and part-of-speech tagging tasks. Furthermore, we introduce the {\it Ottoman Text Corpus (OTC)}, a clean corpus of transliterated historical Turkish texts that spans a wide range of historical periods. Our experimental results demonstrate prominent improvements in the computational analysis of historical Turkish, achieving strong performance on tasks that require understanding of historical linguistic structures—specifically, 90.29\% F1 in named entity recognition, 73.79\% LAS for dependency parsing, and 94.98\% F1 for part-of-speech tagging. They also highlight existing challenges, such as domain adaptation and language variations between time periods. All the resources and models presented are available at \url{https://hf.co/bucolin} to serve as a benchmark for future progress in historical Turkish NLP.   
}

\keywords{computational linguistics, historical Turkish, natural language processing, pre-trained language models, historical language resources
\\\newline
{\small *Corresponding author.}
}



\maketitle

\section{Introduction}
\label{sec1}

Rapid advancements in natural language processing (NLP), particularly due to the success of transformer models, has improved automatic processing in many languages, but the most significant benefits have largely favored widely spoken languages such as English \citep{touvron2023llama}.
Research has largely focused on high-resource languages, while studies on historical languages using transformer models remain scarce \citep{manjavacas-arevalo-fonteyn-2022-non,volk-etal-2024-llm-based}. This scarcity is mainly due to the difficulty in adequately representing many low-resource languages using large data-driven models \citep{lai-etal-2023-chatgpt}. Historical variants of Turkish are among these underrepresented languages from the perspective of current state-of-the-art NLP models and resources.

Ottoman Turkish is the longest-lasting historical variant of the Turkish language, which eventually evolved into modern Turkish. It underwent significant changes in vocabulary and syntax during the six centuries it was in use. Most of the historical Turkish documents preserved in archives were produced during the later centuries of the Ottoman Empire.\footnote{This statement is based on a comparison of the quantities of historical documents from different periods, as preserved in the state archives of the Republic of Türkiye: \url{https://katalog.devletarsivleri.gov.tr/}} Accordingly, this study focuses primarily on Ottoman Turkish from the 18th to the 20th centuries, using texts predominantly from this era. Hereafter, the term {\it historical Turkish} will specifically refer to the Turkish language used during this period.

The digitization efforts of historical documents are rapidly increasing, with the aim of preserving these valuable resources and improving accessibility. The demand for automated analysis and information extraction from these documents is becoming more critical as these historical materials become more available in digital formats. However, meeting this demand is challenging without the necessary and sufficient resources in place. Historical Turkish suffers from lack of annotated data, dictionaries, and linguistic references in contrast to modern languages, which benefit from extensive linguistic resources and corpora. This shortage of resources significantly impedes the development of successful NLP models for historical Turkish.

A possible approach to addressing the lack of resources is to utilize existing language resources for modern Turkish, based on the fact that historical Turkish and modern Turkish are essentially different stages of the same language. However, the transition from historical Turkish to modern Turkish led to significant changes in semantics, vocabulary and grammar \citep{kerslake2021ottoman}. These differences pose obstacles in applying contemporary Turkish NLP techniques directly to historical texts.

Existing tools and resources for processing historical Turkish texts are scarce and often have limited capacity for comprehensive NLP tasks. There have been efforts to develop some datasets and corpora, such as a question-answering dataset \citep{soygazi2021thquad}, a multi-label text classification dataset \citep{gokceoglu2024multi}, and a corpus containing transcripts of Grand National Assembly of Turkey (TBMM) meetings that partially include historical Turkish texts \citep{gungor2018corpus}, but they remain limited in scope and scale.

On the other hand, a few tools developed for historical Turkish mostly address text recognition \citep{bilgin2023printed, DBLP:conf/das/TasdemirTAKSAKY24} and Arabic-to-Latin transliteration \citep{jaf2021machine} tasks, with little to no focus on more complex NLP tasks. To the best of our knowledge, no existing studies have explored advanced NLP tasks for historical Turkish, such as named entity recognition or dependency parsing.

This lack of robust and comprehensive resources highlights a pressing need for the development of advanced tools and datasets to support linguistic processing and computational analysis of historical Turkish texts.

In this pioneering study, we aim to make a comprehensive approach to text analysis of historical Turkish from multiple perspectives and set a foundational starting point for NLP of historical Turkish texts. We provide several resources and models for this purpose. Our contributions are as follows:
\begin{itemize}
    \item The HisTR\footnote{\url{https://huggingface.co/datasets/bucolin/HisTR}} dataset: The first named entity recognition (NER) dataset for historical Turkish, comprising 812 manually annotated sentences from the 17th to the 19th centuries.
    \item The OTA-BOUN\footnote{\url{https://huggingface.co/datasets/bucolin/OTA-BOUN_UD_Treebank}}\textsuperscript{,}\footnote{\url{https://github.com/UniversalDependencies/UD_Ottoman_Turkish-BOUN/tree/dev}} dependency treebank: The first manually annotated dependency treebank for historical Turkish. This treebank includes gold annotation of part-of-speech tags and dependency relations in 514 sentences coming from various literature works. 
    \item Ottoman Text Corpus (OTC)\footnote{\url{https://huggingface.co/datasets/bucolin/OTC-Corpus}}: A clean text corpus spanning from the 15th to the 20th centuries, encompassing texts from various genres and suitable for diverse linguistic purposes.
    \item Transformer-based models trained for dependency parsing, part-of-speech tagging, and named entity recognition tasks, along with their evaluations, to serve as a benchmark for future research in the NLP of historical Turkish.
    \item The release of all the resources and models presented in this study along with the text pre-processing tools for historical Turkish, which have been made publicly available.\footnote{\url{https://huggingface.co/bucolin}}
\end{itemize}

The rest of the paper is organized as follows: Section 2 offers a review of related work on the NLP of historical Turkish. Section 3 introduces the resources developed in this study. Section 4 outlines the NLP tasks that we focus on and explains the models trained for each task. Section 5 describes the experiments conducted and discusses the results. Finally, Section 6 concludes the paper.

\section{Related Work}

The automatic processing of historical documents necessitates preliminary text recognition, typically achieved through Optical Character Recognition (OCR) or Handwritten Text Recognition (HTR), depending on whether the document is printed or handwritten. Although research on NLP for historical Turkish remains scarce, a number of studies have specifically focused on the text recognition of historical Turkish documents.

One of the earliest initiatives for processing historical Turkish documents was the Ottoman Text Archive Project (OTAP), jointly conducted by researchers from Bilkent University and the University of Washington between 2009 and 2012. Although significant progress was made, the project was left unfinished. Publications within the scope of this project include studies on poem categorization \citep{CanCanSahinKalpakli+2013+40+57} and handwriting recognition \citep{DBLP:conf/icpr/CanDCK10}.

Following this initial attempt, a limited number of studies explored OCR for historical Turkish texts, most of which preceded the deep learning era. With the introduction of deep learning techniques, however, a renewed surge of research emerged. For instance, a CNN-LSTM model trained on a combination of synthetic and real data achieved 88.86\% character recognition accuracy and 64\% word recognition accuracy on a test set of printed Naskh line images extracted from 21 pages \citep{DBLP:journals/concurrency/DolekK22}. Similarly, \citet{DBLP:journals/ijdar/Tasdemir23} developed an open-vocabulary system, reporting a character error rate (CER) of 11\% on synthetic data and 16\% on real images. In another study, \citet{DBLP:conf/siu/AydemirAKKD14} proposed an RNN-based system for recognizing handwritten word images from population registration documents, achieving a CER of 12.4\% and a word error rate (WER) of 22.1\% on a small test set of 1,000 unique words. More recently, \citet{DBLP:conf/das/TasdemirTAKSAKY24} introduced an automatic system for transcribing historical Turkish documents into the modern Turkish alphabet, reporting 6.59\% CER and 28.46\% WER on a test set of 6,828 text lines.

As these results indicate, the error rates of OCR and HTR systems remain relatively high. The noise in transcriptions and digitized texts produced by these systems poses an additional challenge for downstream NLP tasks.

Compared to work on other historical languages \citep{fokkens-etal-2014-biographynet,sprugnoli2019novel,lai-etal-2021-event,zilio-etal-2024-nlp}, research on NLP for historical Turkish is extremely limited. A few studies have addressed text classification. One notable example \citep{CanCanSahinKalpakli+2013+40+57} investigated the automatic categorization of Ottoman poems by poet and time period, using Naive Bayes and Support Vector Machines. Their SVM-based approach achieved 88.89\% accuracy for poet classification and 89.19\% for time period classification on a corpus of Ottoman literary texts spanning ten poets and five consecutive centuries. More recently, \citet{gokceoglu2024multi} presented a classification dataset of Ottoman and Russian articles from the late 19th and early 20th centuries. While the dataset includes Ottoman Turkish articles in Perso-Arabic script, it does not provide their Latin-script transcriptions. The authors trained Llama-2 \citep{touvron2023llama}, Falcon \citep{almazrouei2023falconseriesopenlanguage}, and mBERT \citep{devlin-etal-2019-bert} models for multi-level text classification and compared these with a bag-of-words Naive Bayes (BoW-NB) baseline. Interestingly, they found BoW-NB to be competitive with large models in low-resource settings. The best F1 score for first-level single-label classification was 77.65\% with mBERT, while the best second-level single-label classification performance was 83.84\%, achieved by BoW-NB.

Several studies have also attempted the automatic transliteration of Ottoman texts. \citet{kurt2012outline} employed a finite-state machine–based system integrating morphological parsing and word disambiguation. A neural approach using a recurrent neural network (RNN) encoder–decoder architecture for machine translation between modern and historical Turkish was later proposed by \citet{AlNahas}, achieving a BLEU score of 33.8 with a data-driven method for aligning source and target word vectors.

In summary, the majority of existing work on historical Turkish documents has concentrated on text recognition, while only a limited number of studies have explored specific NLP tasks with narrow scope and scale. In this study, we go beyond text recognition by providing gold-standard annotated resources for historical Turkish and establishing baselines for three fundamental NLP tasks: named entity recognition, dependency parsing, and part-of-speech (POS) tagging.

\section{New Resources for Historical Turkish NLP}

Given the extreme scarcity of NLP resources for historical Turkish, our first step is to create and annotate essential datasets and corpora required for automatic text analysis. We manually developed a named entity recognition dataset and a dependency treebank following the Universal Dependencies (UD) annotation scheme for historical Turkish. These datasets were evaluated on named entity recognition, dependency parsing, and part-of-speech tagging tasks, as explained in Section \ref{sec:tasks}. Additionally, we compiled a corpus of transliterated historical Turkish texts that can be utilized for various linguistic purposes. The following subsections introduce these three resources in detail.

\subsection{HisTR: A Historical Turkish NER Dataset}
\label{subsec:HistTR}

We created a NER dataset manually using a subset of sentences from issues of Servet-i Funun  journal (from now on it will be referred to as SF), a historical magazine published between 1896-1901. It covers a wide range of topics including literature, science, daily life and world news.

\begin{figure}[!ht]
\begin{center}

\includegraphics[scale=0.2]{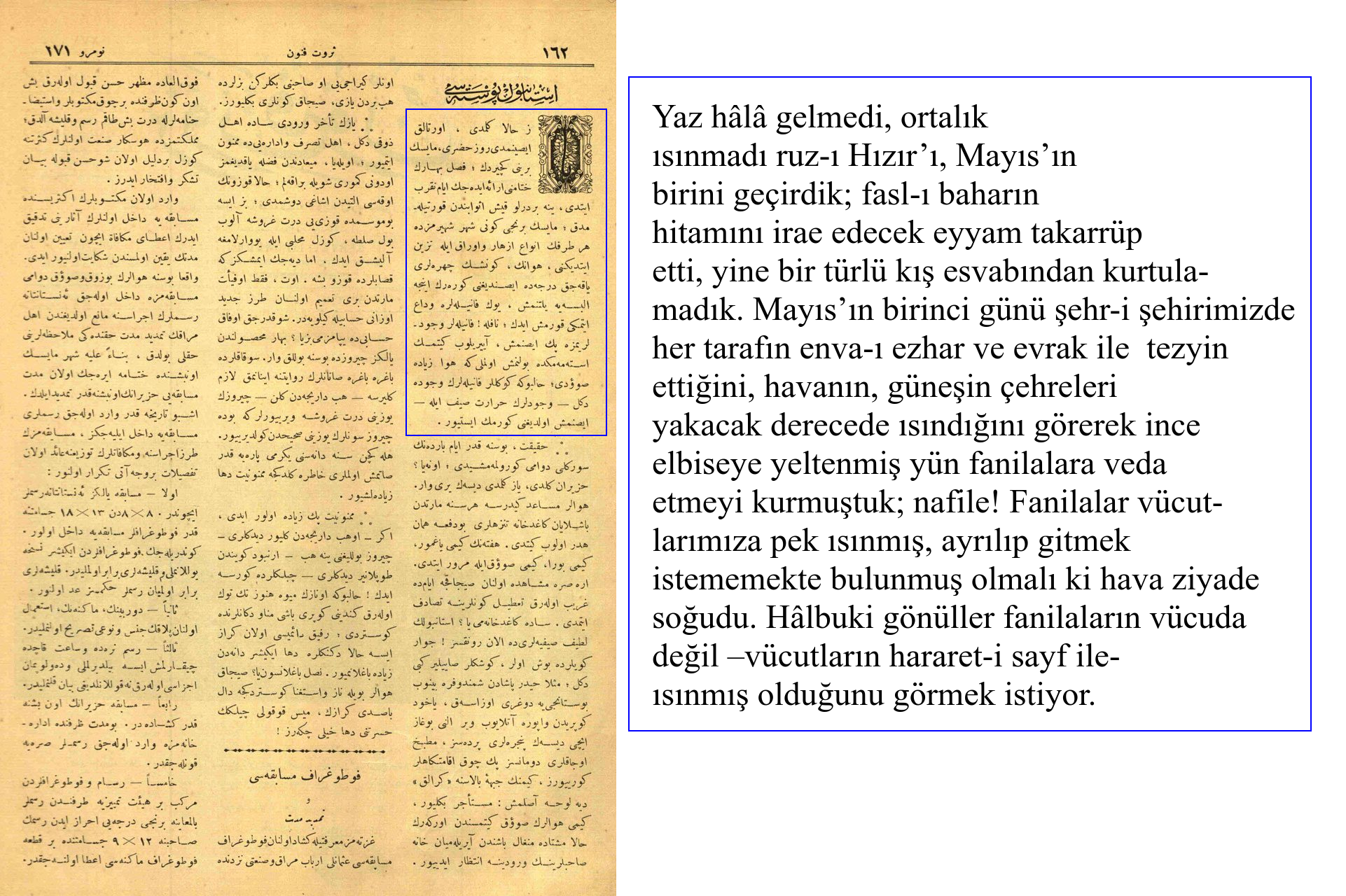} 

\caption{ Transcription of an excerpt from the original document which is written with the Perso-Arabic script.}
\label{fig_SF}
\end{center}
\end{figure}

We used sentences sampled through a research project  conducted between 2016-2019 in Boğaziçi University \citep{SF}. The project aims to give a general view of the periodical by providing original images of pages and transcriptions of some selected sentences. The original script used in the journal is based on the Arabic alphabet while the transcriptions of the sentences are written with the modern Turkish alphabet. Figure \ref{fig_SF} shows an original page from the journal and transcription of an excerpt. 

To create the HisTR dataset, we first annotated 662 sentences from the SF periodicals which were published between the late 19th century and the beginning of the 20th century. Subsequently, to ensure a more objective evaluation of the developed models, we annotated an additional 150 sentences from a collection of {\it Ruznamçe}\footnote{These documents contain records regarding appointments of judges (i.e. {\it qadi}), describing the current and next places of appointments and some other details.} registers from the 17th century. 
The selection criteria of the sentences from {\it Ruznamçe} documents is mainly based on diversity in expressions. The language features of this test set are different from the remaining of the dataset which makes the dataset more challenging in terms of automatic processing. To provide a reference point, the Jensen–Shannon divergence between this {\it Ruznamçe} test set and the training and development sets is 0.86 and 0.85, respectively. In contrast, the divergence between the training and development sets is only 0.47.  

The whole dataset contains 812 sentences, 651 PERSON tags, and 1,010 LOCATION tags.\footnote{We experimented with these two tags only because the number of other tags were not enough to make useful training data.} We use the following definitions for the entity types:

\begin{itemize}
    \item PERSON: People, including fictional
    \item LOCATION: GPE (geo-political entity) and Non-GPE locations including countries, cities, states, mountain ranges and bodies of water.
\end{itemize}

There are various honorifics appearing in the source sentences. We adopted the following approach when dealing with titles, appellations etc. attached to person names:

\begin{itemize}
 \item Titles such as \textit{Doktor ({\it doctor}), Muallim (teacher), Şeyh (sheikh), Madam (madam), Baron (baron), Sör (sir), Matmazel (mademoiselle)} are not included in the PERSON tags.

\item Titles that directly follow personal names, such as \textit{Paşa (pasha), Efendi (sir/master), Bey (mister), Hanım (lady/madam)}, are included within the PERSON tags, because they function as integral parts of personal references rather than separate occupational or honorific titles.

\end{itemize}

The inter-annotator agreement during the manual annotation process was computed using Cohen’s kappa, yielding a score of 0.82. After completion, the annotations were jointly reviewed to resolve any discrepancies.

The data was prepared in CoNLL-2003 format with multiple word entities marked with B- and I- prefixes. Table \ref{tab:ner-dataset} shows the partitions in the dataset together with some statistics. 

\begin{table}[h!]
\caption{\label{tab:ner-dataset}
Partitions in the HisTR dataset}
\centering
\begin{tabular}{lccc}
\hline
  &   \textbf{\# of} & \textbf{PERSON} & \textbf{LOCATION} \\
\textbf{Partition }  &   \textbf{ Sentences} & \textbf{Counts} & \textbf{Counts} \\
\hline
Training set  &   462 & 264 & 584 \\
Development set  &  200 &  122 &  210    \\
{\it Ruznamçe} test set   &  150 & 265 &  216 \\ 
\hline
\bf Total & 812 & 651 & 1,010 \\
\hline
\end{tabular}
\end{table}

\subsection{OTA-BOUN: A Universal Dependencies Treebank for Historical Turkish}

We created the first dependency treebank for historical Turkish as a part of the Universal Dependencies Project \citep{nivre-etal-2017-universal}. 

The annotated sentences were added in two writing styles in the treebank; i) written with the Latin-based Turkish alphabet, and ii) written with the Perso-Arabic alphabet. 

The sentences were sampled from eight texts by
seven distinct writers. All of the texts are from the
literature published between 1880 and 1928. There
are two articles, two excerpts from a historical text, three stories, and one excerpt from a novel in the collection. 

Compiling a treebank from these historically varied texts poses several annotation challenges. Inconsistent spelling arises from shifting orthographic conventions, while specialized or archaic terminology —no longer part of modern usage— complicates morphological analysis. Further complexity comes from the varied stylistic flourishes and personal voices of different authors, each of which can affect syntax, vocabulary, and the treatment of borrowed structures. Inconsistent punctuation also obscures sentence boundaries, requiring extra caution during segmentation. Additionally, the reanalysis of loanwords from Arabic and Persian introduces morphological and semantic ambiguities not always predictable from their roots. Finally, a shortage of standardized references for this historical corpus often necessitates consulting older dictionaries and patchwork resources to accurately parse, interpret, and annotate these texts. A detailed discussion of these and other issues, along with our strategies for addressing them, is provided in Section \ref{challenges}.

\subsubsection{Annotation Scheme}
For the annotation of the historical Turkish treebank, we engaged two expert annotators who are linguists with in-depth knowledge of Turkish grammar, general linguistics and grammatical theory.  Two senior computer scientists possessing significant expertise in both NLP and historical Turkish teamed up with the experts in the annotation task. At this stage, we manually annotated both dependency relations and part-of-speech tags for 514 sentences.

We applied double annotation to a randomly selected subset of 50 sentences. Using Cohen's kappa for dependency labels, we assessed inter-annotator agreement as 0.85. The unlabeled and labeled attachment scores were determined to be 82.20\% and 76.91\%, respectively. The remaining annotations were completed independently by each annotator. Once one annotator completed his/her assigned sections, the annotated sentences were reviewed by the annotation team and any discrepancies were resolved through discussion.

\begin{figure*}[h]
\centering
\fbox{\includegraphics[width=0.95\textwidth]{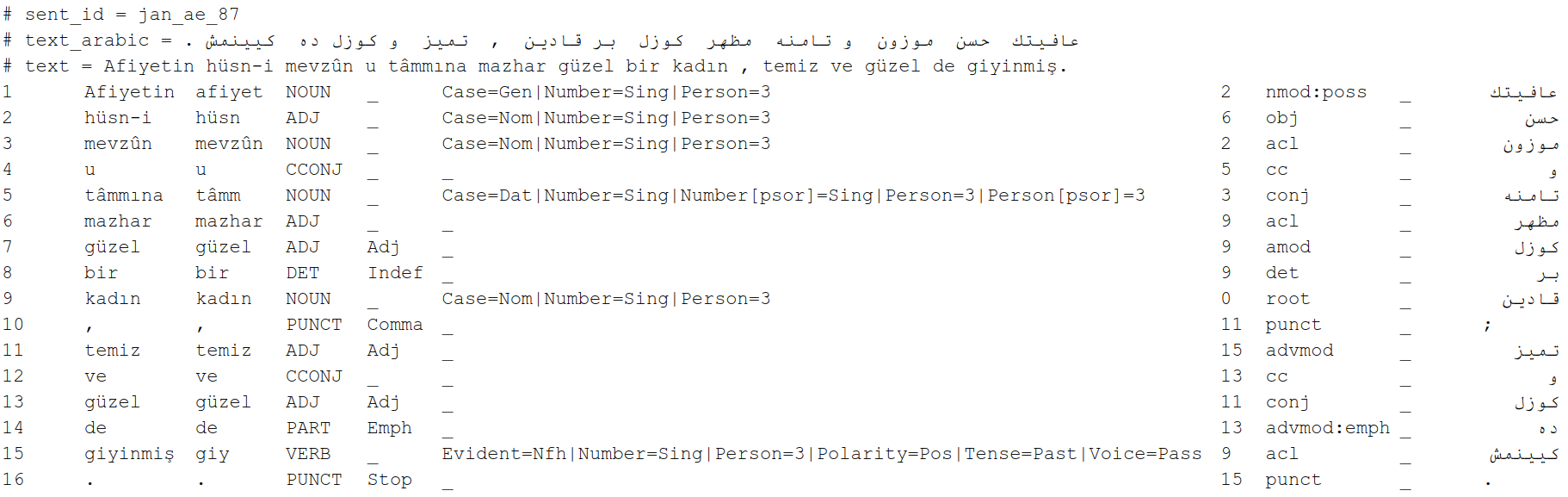}}
\caption{CoNLL-U Representation of an example sentence from our OTA-BOUN historical Turkish treebank.}
\label{fig_ota_boun}
\end{figure*}

We followed the annotation conventions of UD in most cases, but we also consulted the \textit{Suggested UD Guidelines for Turkish}\footnote{\url{https://github.com/boun-tabi/UD_docs/blob/main/_tr/dep/Turkish_deprel_guidelines.pdf}} when needed. In line with the UD framework, the annotated data was saved in CoNLL-U format. Figure \ref{fig_ota_boun} illustrates this format using an annotated sentence from our treebank. We preserve the original Perso-Arabic script version of the data and present the text in both formats. The tokens in Latin script appear in the second column, whereas their Arabic equivalents are displayed in the final column.

\subsubsection{Treebank Statistics}

 The dependency relations, along with their counts and percentages, and basic statistics about the historical Turkish treebank are presented in Tables \ref{basicstats} and \ref{combined-stats}, respectively. Although they reflect partial statistics with respect to the final dataset we plan to annotate, these tables provide valuable insights into historical Turkish, including the average sentence length, the prevalence of specific relations, and other notable linguistic characteristics.

\begin{table}[h]
\caption{\label{basicstats}Some statistics of the OTA-BOUN historical Turkish treebank}
\centering
\begin{tabular}{lc}
\hline
\textbf{Features }  &   \textbf{OTA-BOUN} \\
\hline
Num. of Sentences   &   514  \\
Num. of Tokens  &   8,794    \\
Avg. Token Count Per Sentence   &   17.10   \\
Num. of Unique POS Tags    &   16   \\
Num. of Unique Morphological Features    &    52  \\
Num. of Unique Dependencies    &    40  \\
\hline
\end{tabular}
\end{table}

\begin{table}[h]
\caption{\label{combined-stats}Counts and percentages of dependency relation types in the OTA-BOUN treebank}
\centering
\small
\begin{tabular}{lrr|lrr}
\hline
\textbf{Relation Type} & \textbf{Count} & \textbf{\%} & \textbf{Relation Type} & \textbf{Count} & \textbf{\%} \\
\hline
acl               & 348  & 3.95  & dislocated        & 5    & 0.06  \\
advcl             & 197  & 2.24  & fixed             & 6    & 0.07  \\
advmod            & 396  & 4.49  & flat              & 87   & 0.99  \\
advmod:emph       & 87   & 0.99  & goeswith          & 5    & 0.06  \\
amod              & 620  & 7.04  & iobj              & 26   & 0.30  \\
appos             & 2    & 0.02  & mark              & 27   & 0.31  \\
aux               & 39   & 0.44  & nmod              & 137  & 1.55  \\
case              & 257  & 2.92  & nmod:poss         & 746  & 8.47  \\
cc                & 228  & 2.59  & nsubj             & 507  & 5.75  \\
cc:preconj        & 12   & 0.14  & nsubj:pass        & 22   & 0.25  \\
ccomp             & 120  & 1.36  & nummod            & 57   & 0.65  \\
compound          & 76   & 0.86  & obj               & 557  & 6.32  \\
compound:lvc      & 246  & 2.79  & obl               & 873  & 9.91  \\
compound:redup    & 33   & 0.37  & obl:agent         & 4    & 0.05  \\
conj              & 607  & 6.89  & orphan            & 4    & 0.05  \\
cop               & 48   & 0.54  & parataxis         & 10   & 0.11  \\
csubj             & 42   & 0.48  & punct             & 1207 & 13.70  \\
dep               & 14   & 0.16  & root              & 514  & 5.83  \\
det               & 508  & 5.76  & vocative          & 7    & 0.08 \\
discourse         & 82   & 0.93  & xcomp             & 49   & 0.56  \\
\hline
\end{tabular}
\end{table}

We observe some key differences between the OTA-BOUN historical Turkish treebank and two treebanks for modern Turkish; TR-BOUN and IMST-UD when we compare them quantitatively, as shown in Table \ref{compoundlvc}. 

Numerous Arabic and Persian borrowed words in historical Turkish appear primarily as nominal or adjectival stems that combine with Turkish light verbs (most commonly \textit{etmek}, \textit{olmak}, though others such as \textit{kılmak} and \textit {eylemek} also appear) to form predicates. Hence, relatively frequent occurrences of light verb constructions in texts that contain many borrowings are expected. Table \ref{compoundlvc} presents the frequencies of the light verb compound dependency relation in two modern Turkish UD treebanks, as well as in our historical Turkish treebank. Notably, the frequency of light verb constructions in OTA-BOUN is higher than in the other two modern Turkish treebanks.  

These light verb constructions appear not only as the main predicate (that is, the UD \textit{ root}) but also in subordinate or dependent clauses, showing up in various positions across clausal structures. They allow borrowed lexical items to function as clausal elements, a pattern particularly evident in the frequency of the \textit{acl (adnominal clause)} relation. As shown in Table \ref{compoundlvc}, this relation is notably more frequent in the historical Turkish treebank compared to its modern counterparts.

\begin{table}[h!]
\caption{\label{compoundlvc}Comparison of historical Turkish treebank with the two most frequently used modern Turkish treebanks in terms of token and dependency metrics}
\centering
\small
\begin{tabular}{lcc|c}
\hline
 & \textbf{TR-BOUN} & \textbf{IMST-UD}  & \textbf{OTA-BOUN} \\
\hline
Avg. token count per sent.    & 12.41 & 10.01  & 17.10 \\
conj  (\%)  & 5.66 & 4.96  & 6.89 \\
compound:lvc  (\%) & 1.0  & 0.90   & 2.79 \\
acl     (\%) & 2.78 & 2.64  & 3.95 \\

\hline
\end{tabular}
\end{table}

\begin{figure*}[!ht]
\begin{center}

\includegraphics[scale=0.45]{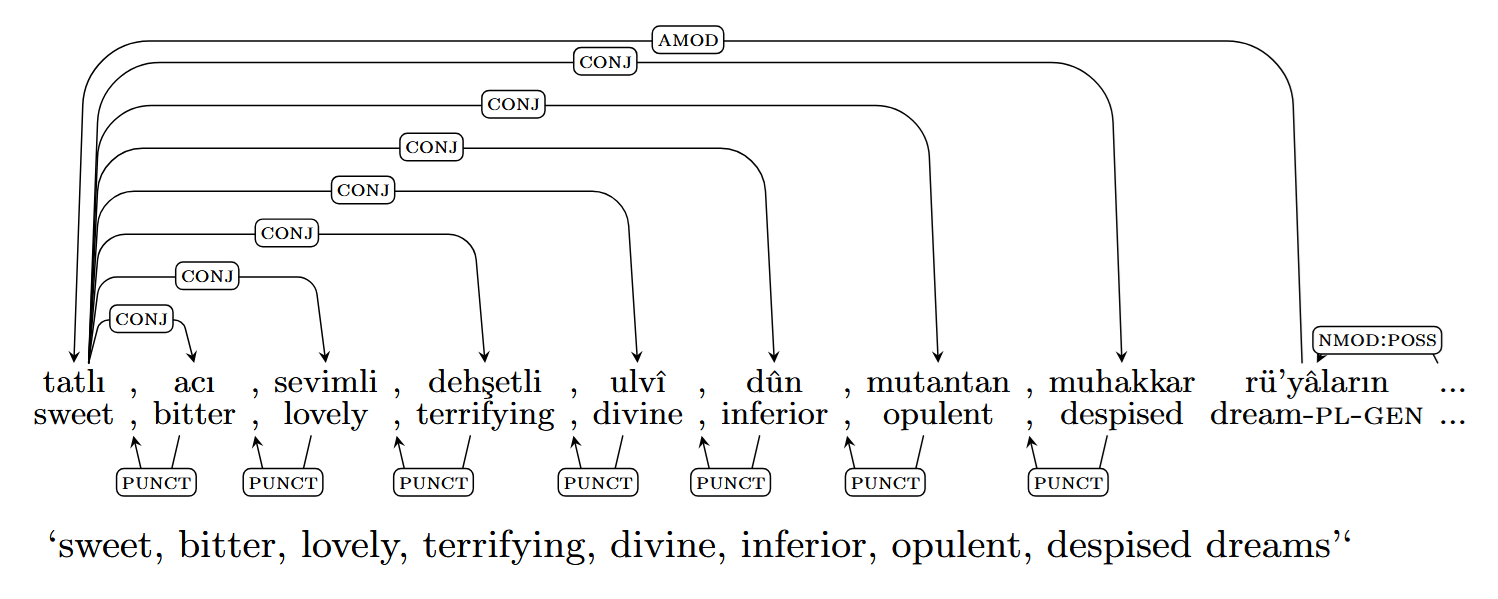} 

\caption{Illustration of frequent coordination in historical Turkish: multiple adjectives (\textit{tatlı, acı, sevimli, dehşetli, ulvî, dûn, mutantan, muhakkar}) are linked by \texttt{conj} and modify the head noun \textit{rü’yâların} “dreams.”}
\label{conjugatzion}
\end{center}
\end{figure*}

Historical Turkish sentences in our dataset tend to be longer on average than those observed in modern Turkish treebanks, although this finding should be considered preliminary given the limited size and literary nature of the sample. As presented in Table \ref{compoundlvc}, the average token count per sentence and the frequency of \textit{conj (conjunct)} relations are higher than those of the modern Turkish UD treebanks. The relatively high frequency of the UD \textit{acl} relation, alongside extensive descriptive modifiers and participial structures, points to a more elaborate writing style. This style often layers nominal phrases with multiple subordinate clauses, which may contribute to the higher observed complexity. An example from our treebank showing numerous \textit{conjunct} relations is provided in Figure \ref{conjugatzion}.

\subsubsection{Challenges in the Syntactic Annotation}
\label{challenges}
Historical Turkish contains numerous loanwords from Arabic whose derivational structure—originally associated with Arabic verb Forms (I, II, V, etc.)—does not always align straightforwardly with their usage in historical Turkish texts. A prime example is the derived word \textit{mütevaggil} \< مُتَوَغِّل >, 
seemingly derived from Arabic Form V \< تَفَعَّل >.
Although the Arabic root generally conveys the sense of “going far or deep into something”, old Ottoman Turkish–English dictionaries attest a nuanced meaning of \textit{mütevaggil} as “someone who occupies himself”, implying an active engagement or absorption in a particular pursuit.

\begin{figure*}[!ht]
\begin{center}

\includegraphics[scale=0.45]{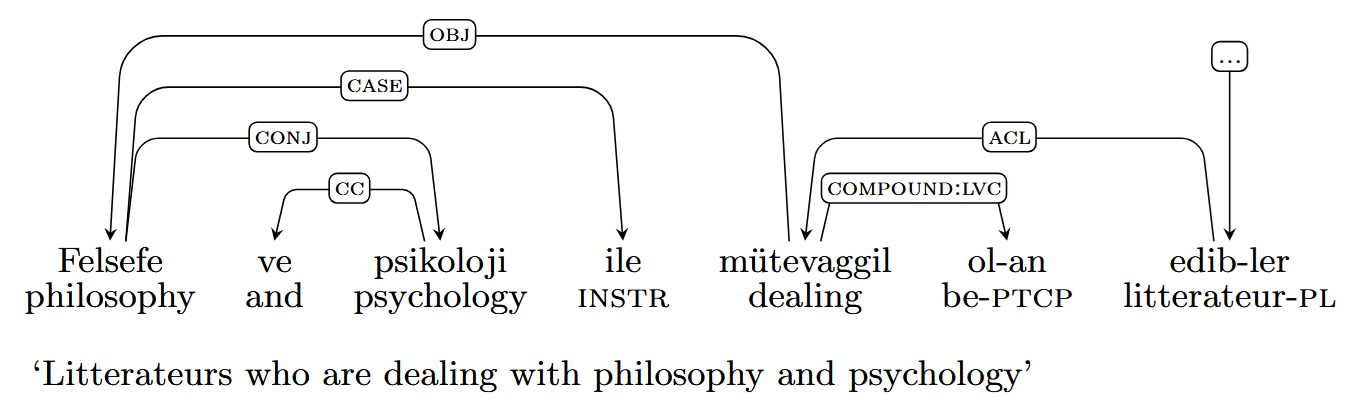} 

\caption{Example of the borrowed participle \textit{mütevaggil} “dealing,” whose argument structure in historical Turkish differs from its Arabic source. The instrumental phrase \textit{felsefe ve psikoloji ile} is annotated as a core complement, highlighting how borrowed forms acquired new syntactic behaviors.}
\label{ex:classifiers3}
\end{center}
\end{figure*}

This difference in meaning highlights a key point: Arabic etymology alone cannot fully explain how loanwords were reanalyzed in historical Turkish. As shown in the example in Figure \ref{ex:classifiers3}, \textit{mütevaggil} here takes an instrumental complement ({\it -ile}), marking the nominal phrase “felsefe ve psikoloji” (philosophy and psychology) as an obligatory argument rather than as an adjunct. Thus, this specific construction encodes “actively dealing with philosophy and psychology”, contrasting with the reflexive or mediopassive nuances often associated with Form V in Arabic.

This subtle shift in meaning demonstrates that even if a borrowed participle is identifiable as Form V in origin, its historical Turkish usage may require different argument structures and yield semantic nuances that diverge from Arabic. Accurately determining the correct annotation—for instance, whether {\it -ile} in such cases functions as a complement rather than an adjunct—often requires consulting dictionaries published during that period, since modern references rarely preserve these now‐archaic or specialized senses. Verifying the original Arabic root and confirming the contemporaneous historical Turkish glosses proved vital to a precise analysis of \textit{mütevaggil} and other historical loanwords.

The necessity of consulting historical dictionaries to confirm these morphological and semantic nuances is not without its own challenges. Many archaic senses have fallen out of use, making them unfamiliar to modern scholars; varying lexicographic sources can conflict in their glosses, complicating efforts to reconcile multiple definitions; orthographic inconsistencies and limited coverage in newer references further obscure older morphological nuances; and older dictionaries often lack explicit grammatical metalanguage, forcing researchers to infer whether a marker like {\it -ile} functions as an argument or adjunct largely from context.

Another notable challenge is that sentences in historical Turkish texts tend to be much longer than those in modern Turkish. As illustrated in Table \ref{compoundlvc}, the average token count per sentence and the relative frequency of conjunctions are notably higher than in previously analyzed modern Turkish UD treebanks. This indicates that the historical Turkish treebank contains substantially longer sentences, with numerous elements linked together, highlighting the complexity and intricacy of sentence structure in historical texts. The length and complexity of historical Turkish sentences led to frequent parsing failures when handling longer structures, complicating the parsing process. Additionally, this complexity significantly slowed down the annotation process, as it became challenging for annotators to fully visualize and interpret the entire sentence structure, requiring more time and effort to ensure accuracy.

The challenges related to the deformation of Turkish morphosyntax, the identification of historical compounds, and obsolete words in the OTA-BOUN historical Turkish treebank have already been discussed in \citet{ozates-etal-2024-dependency}. We refer readers to that study for more details.

\subsection{Ottoman Text Corpus: A Clean Text Corpus of Historical Turkish Documents}

A comprehensive text corpus is essential for many NLP tasks, especially for training language models specifically tailored to historical Turkish. However, there is no such corpus publicly available for historical Turkish NLP yet. 
To address this gap, we developed the first comprehensive transliterated, digital text corpus for historical Turkish: {\it Ottoman Text Corpus (OTC)}.

OTC covers a broad range in time, spanning from the 15th to the 20th century, with a particular focus on the Tanzimat period (1839-1922). During this era, notable improvements were made in systematic use of punctuation, grammar simplification, and the standardization of spelling. There was also a deliberate effort to minimize reliance on foreign loanwords in favor of native Turkish alternatives.

\begin{figure*} 
\centering 
\scalebox{0.9}{\includegraphics[width=1\linewidth]{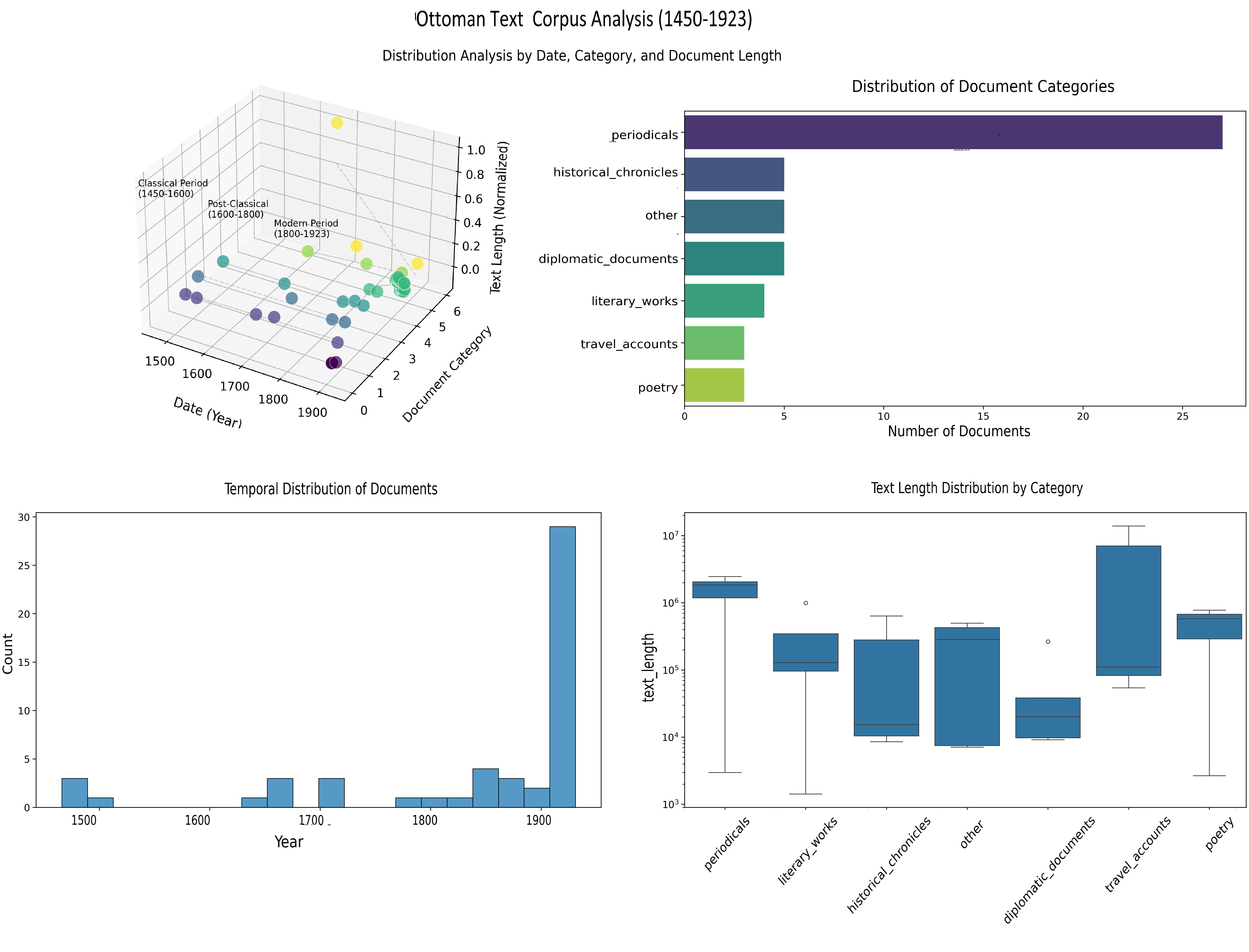}}
    \caption{Visualization of documents in the Ottoman Text Corpus: Each point represents a document, color-coded by topic. This visualization highlights thematic clusters within the corpus and shows how topics are distributed periodically, complementing the map’s representation of vocabulary diversity and topic. }
\label{fig:tsne_ottoman_map} 
\end{figure*}

The initial version \citep{karagoz-etal-2024-towards} of OTC mainly consisted of two Ottoman periodicals, \textit{Sırat-ı Müstakim} and \textit{Sebilürreşşad}. It contained texts from issues published between 1908 to 1923. While this collection serves as the foundation for the corpus, we realized that it is not sufficient for fully capturing the nuanced and localized aspects of historical Turkish needed for advanced NLP utilities. This limitation could lead to overfitting in certain historical applications. To address this issue, we extended OTC by adding a diverse range of literary works such as novels, bibliographies, treaties, newspapers, historical notes, and travelogues from the pre-1908 period. With the latest expansion, OTC now encompasses a total of 11 million tokens. Figure \ref{fig:tsne_ottoman_map} depicts the distribution of documents in the corpus by topic and period.

As can be seen from the figure, the corpus includes text documents from diverse categories and periods. In addition to the texts sourced from periodicals, the OTC corpus also includes texts from various categories such as literary works, diplomatic documents, historical chronicles, and poetry. 

The inclusion of texts spanning from the 15th to the 20th centuries in the corpus offers a unique opportunity to observe the linguistic evolution of historical Turkish over time. However, the broad temporal scope of the corpus poses challenges for adapting modern Turkish PLMs to historical Turkish through continual pre-training. Although the corpus contains approximately 11 million tokens, this size is insufficient for effectively capturing the substantial linguistic variation across historical periods, making a single period-agnostic model impractical. A more appropriate strategy is to isolate subsets representing specific eras and conduct period-focused continual pre-training. Achieving this, however, requires significantly expanding the corpus for each historical period.

Despite these limitations for large-scale model adaptation, the corpus remains valuable for other purposes, such as generating synthetic data and supporting detailed linguistic and diachronic analyses.

\subsubsection{Challenges in Pre-processing}

Creating a clean and well-structured text corpus for historical Turkish poses significant challenges across various dimensions. There is a lack of tools to process historical Turkish documents, particularly those in diverse styles and formats. This scarcity led us to create custom data processing tools designed to facilitate effective modeling in our research.\footnote{We have made these tools and data collections open source at this link: \url{https://github.com/Ottoman-NLP/ottominer-public}.} Our approach includes implementing targeted strategies to tackle the unique linguistic and technical challenges presented by historical Turkish. 

The main technical difficulty stems from PDF extraction artifacts and encoding issues. While our corpus predominantly uses transliterated texts, character representation problems persist, especially with ligatures and complex characters unique to historical Turkish. Table \ref{tab:extraction_problems} illustrates how such issues can result in misrepresentations. Inaccuracies in letter representations not only hinder interpretation but can also produce outputs that deviate significantly from the original text.

\begin{table*}[htbp]
\footnotesize
    \centering
    \caption{Analysis of sample text extraction errors in digital conversion of historical Turkish documents}
   % \begin{tabular}{|>{\raggedright}p{0.22\textwidth}|>{\raggedright}p{0.22\textwidth}|p{0.45\textwidth}|}
   \begin{tabular}{|p{4cm}|p{4cm}|p{4cm}|}
        \hline
        \textbf{Expected Text} & \textbf{Extracted Text} & \textbf{Error Description} \\ 
        \hline
        Dilberün her handesi bin can bağışlar e aşuya & 
        Dil-beruñ her {\color{red}òandesi biñ cÀn baàışlar èÀşıúa} &
        \textbf{Diacritical Encoding Error:}
        Unicode normalization failure in historical Turkish diacritics and characters. The system incorrectly encodes special characters 'ñ' and 'À', resulting in ambiguity.  \\
        \hline
        Bu mutabakatla beraber, keşf edilen eski yazıldığı veçhile Türkçe karşılığı lafzıdır. & 
        Bu mutabakatle beraber, keşf edilen eski ya {\color{red}WU J. ıS i J e Ha Tı Ye Kef Lam Mim Nun} te de yazıldığı veçhile Türkçe karşılığı lafzıdır. &
        \textbf{Script Conversion Error:}
        Critical failure in Arabic-Latin script conversion pipeline. OCR system's inability to properly map Arabic script ligatures to Latin characters due to contextual shape variations.  \\
        \hline
        GÜRİZ yahut GÜRİZGAH:&{\color{red}G Ü R ÎZ} yâhut {\color{red}G Ü R İZ G Â H} : &
        \textbf{Word Segmentation Error:}
        Tokenization algorithm failure in word recognition. Improper word boundary detection caused by missing morphological analysis support. \\
        \hline
        HİSÂB-I CÜMEL: Ebced hisâbının diğer adıdır & {\color{red}HtSÂB-t} C Ü M EL: Ebced hi-sâbının diğer adıdır &
        \textbf{Character Substitution Error:}
        Systematic misclassification of Turkish 'İ' and 'I' characters as 't'.  \\
        \hline
    \end{tabular}
    \label{tab:extraction_problems}
\end{table*}

We employ regex-based static rules following \citet{karagoz-etal-2024-towards} to address these technical artifacts. While effective as a baseline, our ongoing research highlights the need for a more adaptive approach to achieve comprehensive solutions in the future. The substantial variance in error types, coupled with the inherent complexity of historical Turkish orthography, underscores the limitations of static extraction rules. Currently, to ensure that the extracted texts are accurate and consistent, we employ a manual cleaning step as the final stage of the process.

\section{Tasks and Models}
\label{sec:tasks}
We trained transformer-based models for named entity recognition, dependency parsing, and part-of-speech tagging tasks specific to historical Turkish. This section provides an overview of these tasks and the models employed. 

\subsection{Named Entity Recognition}
Named entity recognition is an NLP technique that involves identifying and classifying named entities such as people, locations, and organizations within text data. The goal of NER is to extract meaningful information from unstructured text to help with further analysis or understanding.

Applying NER to historical texts is a \textit{relatively new domain} in NER studies \citep{Ehrmann:2020}. Actually, NER plays a crucial role in document indexing, keyword searching, and information extraction from historical documents.

There are many issues to be considered in the NER process: ambiguous words, variations due to abbreviations, and out-of-vocabulary words pose challenges during NER tasks. Most of the time, NER models trained on general text data may not perform well in domain-specific or specialized text, such as medical, legal, or scientific documents. As a solution, such models are adapted to specific domains by fine-tuning them, which requires domain-specific training data. However, generating labeled datasets is a labor-intensive and error-prone task.

Recognition of named entities in historical texts is particularly difficult due to several additional reasons. One problem is the evolution of language over time, resulting in differences in spelling, grammar, and vocabulary. Similarly, the changing nature of real-world entities like political borders, city names, administrative divisions poses further challenges. Finally, in case of using an OCR system for transcription generation, recognition errors  make NER of historical texts a more difficult task. 

Creating a single NER model that covers all historical periods is a complex task. A common approach is to generate specialized NER models and datasets tailored to historical texts. Actually, manual curation and annotation of historical data by domain experts can improve the accuracy of NER models when dealing with historical texts \citep{Ehrmann:2020}.

The creation of the HisTR NER dataset enables us to explore the named entity recognition task for historical Turkish texts. We employ and fine-tune transformer-based language models that have demonstrated groundbreaking performance across a wide range of NLP tasks. Through a series of experiments, we aim to identify the most effective NER model tailored specifically for historical Turkish. 

\begin{figure*}[!ht]
\begin{center}

\includegraphics[scale=0.6]{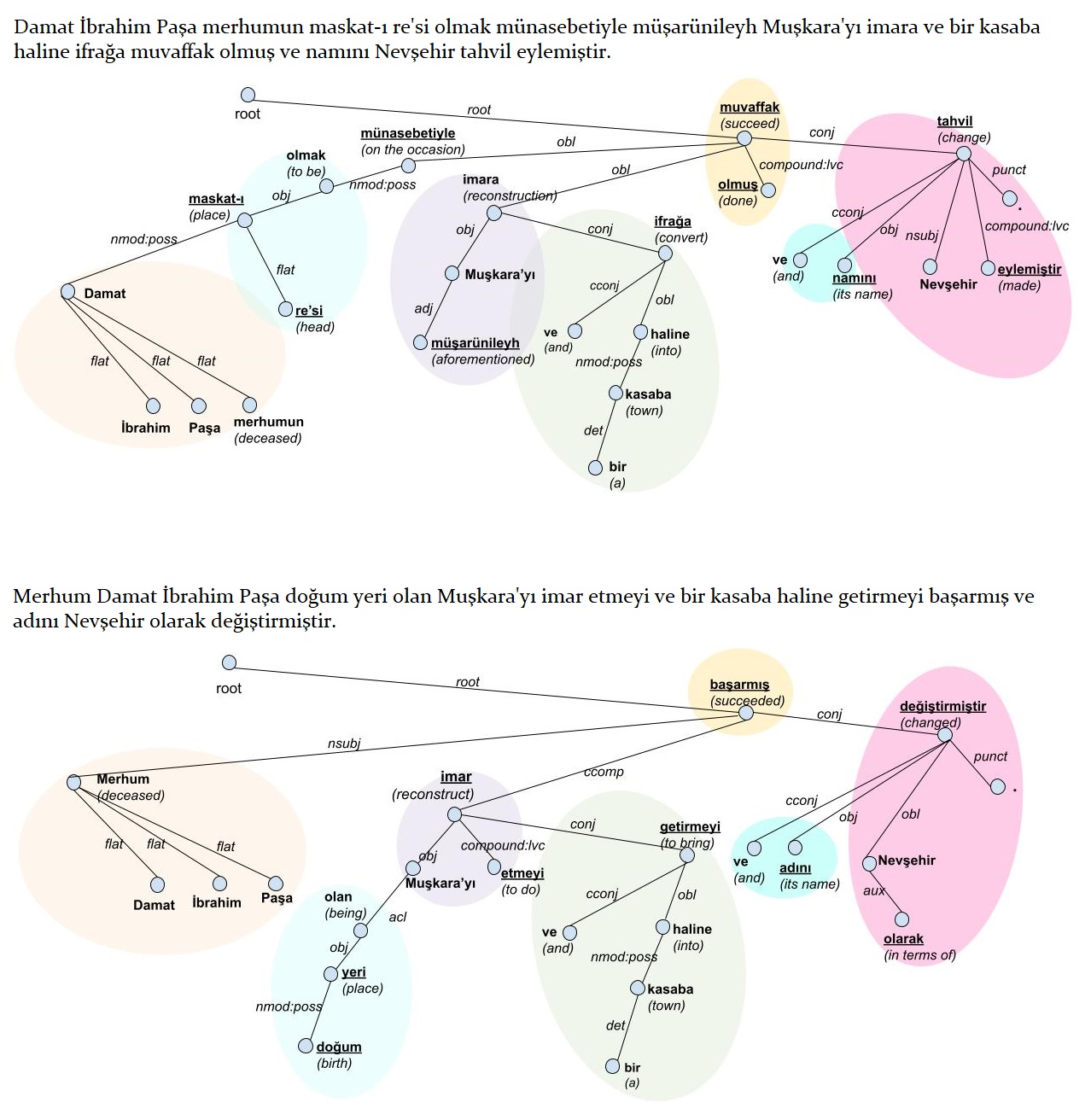}

\caption{ Dependency tree representations of a historical Turkish sentence (above) and its rewritten version in modern Turkish (below). The highlighted portions enclosed in colored circles indicate corresponding segments in the sentences. English translations of words are provided in italic within parentheses. Words of a sentence that do not exist in the other sentence are underlined in the figure. {\textbf{English translation of the sentence:}} "The late Damat İbrahim Paşa managed to develop Muşkara, his birthplace, and turn it into a town, and changed its name to Nevşehir."}
\label{fig_deptree}
\end{center}
\end{figure*}

\subsection{Dependency Parsing and Part-of-Speech Tagging}

Dependency parsing is a core NLP task that analyzes the grammatical structure of a sentence to establish relationships between words. It identifies how words are connected to each other, forming a tree-like structure called a dependency tree. Since a dependency tree of a sentence consists of a set of dependent-head pairs, the order of
words in the sentence is not important and does not change construction of the tree. This flexibility makes dependency tree representations particularly well-suited for languages with free word order, such as Turkish, compared to constituency tree representations \citep{ozates2022dep}.

Another fundamental NLP task that is closely related to dependency parsing is POS tagging. POS tagging is a sequential labeling task that involves assigning a part of speech (e.g., noun, verb, adjective) to each word in a sentence, which helps determine the role of each word within the sentence. Accurate POS tagging can greatly improve the performance of dependency parsing, as understanding the syntactic roles of words provides essential context for establishing correct head-dependent relationships. For this reason, these two tasks have often been studied together in the literature \citep{dozat-etal-2017-stanfords, vilares2020parsing, ozatecs2022hybrid}, leading to the development of joint models that exhibit superior performance \citep{zhang-etal-2015-randomized, yang2017joint,zhou2020pos}. These tasks together contribute to a deeper understanding of sentence structure through syntactic analysis and are essential for a variety of advanced NLP applications, including relation extraction \citep{tian-etal-2021-dependency}, coreference resolution \citep{meng2023rgat}, and sentiment analysis \citep{dashtipour2020hybrid}.

Syntactic analysis of historical languages, however, is not a well-explored topic. There have been a few studies focusing on this problem \citep{keersmaekers2020creating, grobol2022bertrade} and most of them have not reached the desired success levels due to the challenges posed by historical texts. 

Historical texts typically feature outdated or obsolete vocabulary and grammatical constructions, which complicate the direct application of contemporary models to them. Figure \ref{fig_deptree} illustrates the dependency trees of an example sentence in historical Turkish and its rewritten version in modern Turkish. The differences in grammatical structure between the two sentences are evident in their dependency trees. For instance, the way the subject of the sentence ({\it Damat İbrahim Paşa}) connects to the verb differs: in the original sentence, it is linked via an {\textit{oblique}} relation through one of its parent nodes, whereas in the modern version, it is connected directly via a {\textit{nominal subject}} relation. Another notable difference is how the subject’s birthplace ({\it Muşkara}) is indicated. In the original sentence, this relationship is established indirectly through a connection at the root word, while in the modern version, it is conveyed more directly using a clausal modifier.
These structural differences make the syntactic analysis of historical Turkish texts more challenging compared to modern Turkish texts.

We employ transformer-based models for dependency parsing and POS tagging tasks, leveraging their ability to capture long-range dependencies and complex syntactic patterns. Through our experiments, we examine their adaptability to the unique linguistic challenges posed by historical texts, such as non-standard spellings, unknown vocabulary, and grammatical variations. 

\subsection{Pre-trained Language Models}

Given the limited size of our training data, we chose to pursue leveraging the advanced capabilities of widely used PLMs, rather than attempting to train a new model from scratch with a minimal amount of labeled data. To identify pre-trained models potentially suited to texts written in historical Turkish, we compared the performance of a number of PLMs previously utilized for the NER and POS tagging tasks on Modern Turkish \citep{Schweter:2020}. Among these models, BERTurk, a Turkish language model utilizing the BERT architecture and pre-trained on Turkish text, achieved solid performance consistently for the NER and POS-tagging tasks on various Turkish datasets. In fact, BERTurk has demonstrated
state-of-the-art performance across a wide range
of natural language understanding tasks in Turkish \citep{uzunoglu-sahin-2023-benchmarking,NAJAFI2024124737}  

Additionally, TURNA \citep{uludogan-etal-2024-turna}, a recently proposed T5-based encoder-decoder Turkish language model, built on the Unifying Language Learning (UL2) framework \citep{tay2022ul2}, was introduced for both understanding and generation tasks in Turkish. Both BERTurk and TURNA demonstrated similar performance in several NLP tasks \citep{uludogan-etal-2024-turna}. Hence, we utilized both models for NER tagging of historical Turkish texts.

Since historical Turkish shares a common vocabulary with modern Turkish, along with a significant number of words that have become obsolete or have evolved to be unrecognizable in contemporary usage, we hypothesize that a multilingual PLM may also be suitable for our datasets. We consider multilingual BERT (mBERT) \citep{devlin-etal-2019-bert} and XLM-R \citep{conneau2019unsupervised} architectures for this aim and observe that mBERT performs slightly better than XLM-R for named entity recognition of modern Turkish in our preliminary studies. Hence, we opted to utilize mBERT as the multilingual model for our tasks.  

In addition to PLMs, we also carried out preliminary NER experiments with large language models, including LoRA-based fine-tuning of Llama 3.2 and several prompt-based approaches. However, these LLM-based methods yielded notably lower performance on historical Turkish NER, largely due to their limited suitability for low-resource, domain-specific settings and their difficulty adapting effectively with the small amount of supervised data available. Due to their unsatisfactory and unstable results compared to fully fine-tuned BERTurk, TURNA, and mBERT, we did not include LLM-based approaches in the main experimental setup.

\subsubsection{Model Details}

\textbf{BERTurk} is a language model specifically tailored for the Turkish language, built upon the architecture of Bidirectional Encoder Representations from Transformers (BERT). As in the architecture of the original BERT base model, BERTurk has 12 transformer layers. Each transformer layer consists of 12 attention heads and the number of hidden units is 768. A total of 110 million parameters are fine-tuned during the pre-training phase on a large corpus of Turkish text data, allowing the model to learn contextual representations that capture intricate syntactic and semantic relationships within the language. This architecture, similar to other BERT-based models, has been highly successful in various natural language understanding tasks and has become a cornerstone for Turkish language processing research and applications.

\textbf{Multilingual BERT (mBERT)} is a variant of the BERT architecture that has been pre-trained on text data from 104 languages, making it a multilingual language model. mBERT follows the same general architecture as BERT, typically with a base model consisting of 12 transformer layers. Each transformer layer includes a certain number of attention heads (usually 12) and hidden units (often 768). mBERT is pre-trained on a vast and diverse corpus of text from numerous languages. The goal is to expose the model to a wide range of languages and linguistic structures, enabling it to learn multilingual representations that capture similarities and differences between languages.
The mBERT model is particularly valuable for multilingual natural language processing tasks, as it can be fine-tuned for various downstream applications in different languages without the need for language-specific pre-trained models. Its ability to understand and generate text in multiple languages has made it a useful tool in cross-lingual and multilingual applications.

\textbf{TURNA} is a Turkish language model built on the T5 architecture \citep{raffel2020exploring}, which employs an encoder-decoder transformer framework. TURNA has 36 encoder and decoder layers, with each layer containing 16 attention heads. The model’s token embeddings are 1,024 dimensional. Its multi-layer perceptron layers have 2,816 hidden dimensions. TURNA includes 1.1 billion parameters, making it significantly larger than traditional BERT-based models. Pre-trained using the UL2 framework, TURNA incorporates multiple training objectives, including span corruption and autoregressive tasks, enabling it to perform both understanding and generation tasks.

\subsubsection{Fine-tuning}
We leveraged each PLM by fine-tuning its pre-trained weights on our specific tasks using the limited training data available in the corresponding historical Turkish datasets. Specifically, we utilize HisTR for NER tagging and OTA-BOUN for dependency parsing and POS tagging. Additionally, we incorporated the PLMs fine-tuned on extensive datasets for modern Turkish, evaluating their performance on the historical Turkish test sets to examine cross-domain transferability.

\section{Experimentation}

We conducted experiments to evaluate the HisTR dataset on NER tagging, the OTA-BOUN treebank on dependency parsing and POS tagging tasks. 

\subsection{NER Experiments}

\subsubsection{Data}
 We used HistTR which is introduced in Section \ref{subsec:HistTR} for fine-tuning the PLMs on the NER tagging. The dataset is randomly split into training and test subsets. The training set contains 11,852 tokens (462 sentences) with a total number of 584 LOCATION and 264 PERSON entities. There are 5,101 tokens (200 sentences) with a total number of 210 LOCATION and 122 PERSON entities in the development set. In the out-of-domain {\it Ruznamçe} test set, there are 6,386 tokens (150 sentences) with 216 LOCATION and 265 PERSON entities.

We also used MilliyetNER \citep{Tur:2003} and WikiANN \citep{rahimi-etal-2019-massively} datasets as additional labeled data in the fine-tuning process. MilliyetNER was collected from news articles and manually annotated with PERSON, LOCATION, and ORGANIZATION entity types. It can be considered a large dataset with almost 500 K tokens and stands as one of the most frequently used NER dataset in Turkish.

WikiANN is a multilingual NER dataset consisting of Wikipedia articles from 176 languages, automatically annotated with the entity types of LOCATION, PERSON, and ORGANIZATION.

\subsubsection{Experimental Settings}

 We used Huggingface’s Trainer API for fine-tuning the BERTurk and mBERT models. We used the cased versions of both models. We fine-tuned the models using the Adam optimizer with a learning rate of 5e-5 and a batch size of 32. Training continues until convergence with a maximum epoch number of 20. 
 
 All of the BERT-based experiments were conducted on the Google Cloud platform Colab with 12.7GB RAM and one Tesla T4 GPU. TURNA experiments required the use of mixed precision at FP16 and a smaller batch size on an A100 GPU with 40 GB of RAM.

\subsubsection{Results}
We conducted a series of experiments using BERTurk, mBERT, and TURNA models. Table \ref{results-ner} shows the performance of the models in terms of precision, recall and F1 scores as well as descriptions to clarify the model names. For each model, we run the experiments three times and report the mean scores with their standard deviations.

\begin{table*}[!ht]
\caption{\label{results-ner}
The overall performance of BERTurk, mBERT, and TURNA NER models on the {\it in-domain} development and {\it out-of-domain} test sets of the HisTR dataset when using different combinations of fine-tuning sets
}
\centering
\small
\addtolength{\tabcolsep}{-0.1em}
\begin{tabular}{lcccccc}
\bf Model Descriptions\\
\hline
\texttt{BERTurk+Milliyet} & \multicolumn{6}{l}{\tt BERTurk fine-tuned only using MilliyetNER, a large NER}\\
 &\multicolumn{6}{l}{\tt dataset for modern Turkish.}\\
 \hline
 \texttt{BERTurk} & \multicolumn{6}{l}{\tt BERTurk+MilliyetNER further fine-tuned using HisTR, the small}\\
 \hspace{12pt}\texttt{+Milliyet+HisTR}&\multicolumn{6}{l}{\tt dataset for historical Turkish.}\\
 \hline
 \texttt{BERTurk+HisTR} & \multicolumn{6}{l}{\tt BERTurk fine-tuned only using HisTR.}\\
 \hline
 \texttt{mBERT} & \multicolumn{6}{l}{\tt mBERT fine-tuned on WikiANN, a large multilingual NER dataset,}\\
 \texttt{+WikiANN+HisTR}&\multicolumn{6}{l}{\tt and further fine-tuned using HisTR.}\\
 \hline
  \texttt{mBERT+HisTR} & \multicolumn{6}{l}{\tt mBERT fine-tuned only using HisTR.}\\
 \hline
 \texttt{TURNA} & \multicolumn{6}{l}{\tt TURNA fine-tuned on MilliyetNER and further fine-tuned using}\\
\hspace{12pt}\texttt{+Milliyet+HisTR} &\multicolumn{6}{l}{\tt HisTR.}\\
 \hline
\\
\bf Model Performance\\
\hline
& \multicolumn{3}{c}{HisTR Development Set} & \multicolumn{3}{c}{{\it Ruznamçe} Test Set}\\
\textbf{Name} &  \textbf{Prec.} & \textbf{Recall} & \textbf{F1} & \textbf{Prec.} & \textbf{Recall} & \textbf{F1}\\
\hline
\texttt{BERTurk+Milliyet} & $70.15_{\pm0.56}$  & $74.09_{\pm0.52}$ & $72.06_{\pm0.54}$ & $53.84_{\pm2.71}$ & $61.95_{\pm1.57}$ & $57.58_{\pm1.83}$ \\
\texttt{BERTurk} & &&&&& \\
\hspace{12pt}\texttt{+Milliyet+HisTR} & \bf \multirow{-2}{*}{$88.82_{\pm1.23}$} & \bf \multirow{-2}{*}{$91.32_{\pm1.24}$} & \bf \multirow{-2}{*}{$90.05_{\pm1.20}$} & \bf \multirow{-2}{*}{$55.61_{\pm3.85}$} & \bf \multirow{-2}{*}{$\mathbf{62.48_{\pm1.36}}$} & \bf \multirow{-2}{*}{$\mathbf{58.84_{\pm2.72}}$}\\
\texttt{BERTurk+HisTR} &$\mathbf{89.15_{\pm1.33}}$ & $\mathbf{91.52_{\pm1.28}}$ & $\mathbf{90.29_{\pm0.99}}$ & $54.26_{\pm0.97}$ & $56.32_{\pm4.71}$ & $55.20_{\pm2.39}$ \\
\hline
\texttt{mBERT} & &&&&& \\
\hspace{12pt}\texttt{+WikiANN+HisTR} & \multirow{-2}{*}{$79.92_{\pm2.67}$} & \multirow{-2}{*}{$84.93_{\pm1.61}$} & \multirow{-2}{*}{$82.34_{\pm2.13}$} & \multirow{-2}{*}{$41.17_{\pm1.81}$} & \multirow{-2}{*}{$41.93_{\pm2.22}$} & \multirow{-2}{*}{$41.49_{\pm0.78}$} \\
\texttt{mBERT+HisTR} & $84.07_{\pm0.49}$ & $87.01_{\pm0.88}$ & $85.52_{\pm0.66}$ & $46.03_{\pm3.54}$ & $43.50_{\pm1.47}$ & $44.66_{\pm1.82}$ \\
\hline
\texttt{TURNA} &&&&&&\\
\hspace{12pt}\texttt{+Milliyet+HisTR} &  \multirow{-2}{*}{$78.17_{\pm2.06}$} & \multirow{-2}{*}{$80.77_{\pm1.08}$} & \multirow{-2}{*}{$79.45_{\pm1.58}$} & \multirow{-2}{*}{$\mathbf{57.61_{\pm0.29}}$} & \multirow{-2}{*}{$41.58_{\pm0.87}$} & \multirow{-2}{*}{$48.30_{\pm0.48}$}  \\
\hline
\end{tabular}

\end{table*}

\paragraph{\bf Performance Comparison on the {\it in-domain} Development Set:}

We observe that BERTurk outperforms both mBERT and TURNA in all settings for NER tagging of historical Turkish. This suggests that although historical Turkish differs from modern Turkish, the linguistic similarities may allow BERTurk to transfer its learned representations more effectively than mBERT, which has a broader but less specialized knowledge base.

When we look at the performance of BERTurk, we see that fine-tuning BERTurk using labeled data that include texts written only in modern Turkish (the first row) does not yield good results (F1=72.06\%), even if the labeled data (i.e., MilliyetNER) is quite extensive as in our case. Further fine-tuning the model with the training set of our HisTR dataset (the second row) improved the performance by a large margin (F1=90.05\%). However, note that when fine-tuning BERTurk with only the modest training set of the HisTR dataset and not using the large MilliyetNER dataset (the third row), the performance remains similar (F1=90.29\%). This finding suggests that NER of historical Turkish poses different challenges that do not exist in NER of modern Turkish.  

Upon examining the experimental results conducted with mBERT, we observe a similar pattern in the usage of historical Turkish data. However, inclusion of an extensive multilingual labeled dataset (WikiANN), which has labeled data in modern Turkish besides other 175 languages, during the fine-tuning process adversely impacts the model's performance. Hence, we can conclude that fine-tuning mBERT with multilingual data does not have a positive effect in our task according to the overall scores. 

The final row shows the performance of the T5-based TURNA model fine-tuned on MilliyetNER and then with HisTR. The performance of TURNA on the HisTR development set is the worst one between the models fine-tuned with HisTR. We should note that, during fine-tuning of TURNA, the training loss remained higher than the validation loss, even after convergence. We attribute this to TURNA's large number of parameters, which require more data for effective fine-tuning than what is available in the HisTR dataset. Notably, this issue was not observed with the BERT models, which have moderate size and performed well with the same dataset size.

\paragraph{\bf Performance Comparison on the {\it out-of-domain} Test Set:}

The {\it Ruznamçe} test set is highly divergent linguistically from the rest of the HisTR dataset, as elaborated in Section~\ref{subsec:HistTR}. From Table~\ref{results-ner}, our first observation is the dramatic performance drop across all models. The best-performing model (\texttt{BERTurk+Milliyet+HisTR}) achieved an F1 score of 58.84 on this test set, while fine-tuning BERTurk only on MilliyetNER yielded a very similar performance (F1 = 57.58). The substantial performance improvement observed in the development set when incorporating HisTR into fine-tuning is not reflected when the test data differs significantly from the sentences in the HisTR training portion. This notable gap can be attributed to the temporal and contextual variations inherent in the datasets, which originate from distinct historical periods. While the sentences in the HisTR dataset are sourced from a 19th-century periodical, the {\it Ruznamçe} test set consists of 17th-century legal documents.

The worst-performing model in the development set, TURNA, outperforms the best mBERT configuration by approximately 4 points on the out-of-domain {\it Ruznamçe} test set, although it still lags behind all BERTurk configurations. These results suggest that TURNA’s domain adaptation capability is stronger than that of mBERT. In addition, TURNA exhibits the highest precision among all models, but it has the worst recall. 

From these results, we can claim that the mediocre performance of all models on the {\it Ruznamçe} test set underscores the need for new methods and the development of more comprehensive datasets to enable effective NER tagging of historical Turkish texts across diverse domains and time periods.

%This raises the question of whether the observed low recall stems from the strict matching criteria employed in the evaluation scheme.

%When comparing the performance of the models on the in-domain development set and the out-of-domain {\it Ruznamçe} test set, a notable disparity emerges. The best-performing model achieves an F1 score of only 61.91 on the {\it Ruznamçe} test set, compared to 91.21 on the development set—a difference of nearly 30 points. 

\subsection{Dependency Parsing and POS Tagging Experiments}

\subsubsection{Data}

The models were trained and evaluated on both the OTA-BOUN treebank, the first and only dependency treebank for historical Turkish and the Turkish BOUN treebank \citep{turk2022resources}, a large dependency treebank for modern Turkish. Both treebanks are in UD format, containing manual annotation of universal part-of-speech tags and dependency relations. 

We adhere to the original partitioning of the treebanks, where the OTA-BOUN treebank consists of only 114 sentences in the training set and 400 sentences in the test set. In contrast, the Turkish BOUN treebank features a much larger dataset, with 7,803 sentences in the training set and 979 sentences in both the development and test sets.

\subsubsection{Experimental Settings}

For the dependency parsing and POS tagging experiments, we utilized the STEPS parser \citep{grunewald-etal-2021-applying}. STEPS is a graph-based dependency parser built on the well-known biaffine classifier approach \citep{dozat-etal-2017-stanfords} but incorporates transformer-based encoders in its internal architecture. For the transformer encoders, we experimented with BERTurk and mBERT.

To configure STEPS for dependency parsing, we adopted the following settings to optimize its performance: the arc scorer was assigned a hidden size of 768 with a dropout rate of 0.33, while the label scorer used a hidden size of 256 with the same dropout rate. The embeddings processor was configured with hidden, attention, and output dropout rates set to 0.2, 0.2, and 0.5, respectively, complemented by a token mask probability of 0.15. For POS tagging, we employed a sequence tagger featuring an input dropout rate of 0.2, while retaining the same embedding processor configuration as used for dependency parsing. Both tasks leveraged the Adam optimizer with a learning rate of 4e-5, combined with a square root learning rate schedule spanning 400 steps.

Experiments of both tasks were conducted on the Google Cloud platform using Colab, with 16 GB of RAM and a Tesla T4 GPU. For both tasks, we used a batch size of 32 and implemented early stopping with a patience of 15 epochs, allowing a maximum of 300 epochs.
 
\subsubsection{Results}

Table \ref{results-all-depparse} presents the results of the experiments conducted using STEPS parser with BERTurk and mBERT models. A description of the model names is provided in the upper part of the table for clarification. We perform three runs for each model and report the average scores with corresponding standard deviations.

\begin{table*}[!ht]
\small
\caption{\label{results-all-depparse}
The overall performance of BERTurk- and mBERT-based models on the test sets of the OTA-BOUN and TR-BOUN treebanks, presented for different combinations of fine-tuning sets. UAS and LAS represent unlabeled and labeled attachment scores, respectively, which are used to evaluate the models' performance in constructing dependency relations. UPOS F1 refers to the F1 score of the sequence tagger models in predicting the universal POS tags of words in the corresponding test sets.
}
\centering
%{\fontsize{7.5}{10.5}\selectfont
\addtolength{\tabcolsep}{-0.15em}
\begin{tabular}{lccccccc}  % Changed to 7 columns
\bf Model Descriptions\\
\hline
$\text{STEPS}_{\text{BERTurk}}$\texttt{+TR\_BOUN} & \multicolumn{7}{l}{\tt STEPS parser with the BERTurk encoder, fine-tuned only using}\\
 &\multicolumn{7}{l}{\tt TR\_BOUN, a large dependency treebank for modern Turkish.}\\
 \hline
 $\text{STEPS}_{\text{BERTurk}}$ & \multicolumn{7}{l}{$\text{STEPS}_{\text{BERTurk}}$\texttt{+TR\_BOUN} \tt further fine-tuned using OTA\_BOUN,}\\
 \hspace{12pt}\texttt{+TR\_BOUN}\texttt{+OTA\_BOUN}&\multicolumn{7}{l}{\tt a small treebank for historical Turkish.}\\
 \hline
 $\text{STEPS}_{\text{BERTurk}}$\texttt{+OTA\_BOUN} & \multicolumn{7}{l}{\tt STEPS parser with the BERTurk encoder, fine-tuned only using}\\
 &\multicolumn{7}{l}{\tt OTA\_BOUN}\\
 \hline
 $\text{STEPS}_{\text{mBERT}}$\texttt{+TR\_BOUN} & \multicolumn{7}{l}{\tt STEPS parser with the BERTurk encoder, fine-tuned only using}\\
 &\multicolumn{7}{l}{\tt TR\_BOUN.}\\
 \hline
 $\text{STEPS}_{\text{mBERT}}$ & \multicolumn{7}{l}{$\text{STEPS}_{\text{mBERT}}$\texttt{+TR\_BOUN} \tt further fine-tuned using OTA\_BOUN.}\\
 \hspace{12pt}\texttt{+TR\_BOUN}\texttt{+OTA\_BOUN}&&&&&&& \\
 \hline
 $\text{STEPS}_{\text{mBERT}}$\texttt{+OTA\_BOUN} & \multicolumn{7}{l}{\tt STEPS parser with the mBERT encoder, fine-tuned only using}\\
 &\multicolumn{7}{l}{\tt OTA\_BOUN.}\\
 \hline
\\
\bf Model Performance\\
\hline
& & \multicolumn{3}{c}{OTA-BOUN Test Set} & \multicolumn{3}{c}{TR-BOUN Test Set} \\ 
& & \multicolumn{3}{c}{\it (Historical Turkish)} & \multicolumn{3}{c}{\it (Modern Turkish)} \\ 
\textbf{Name} & \bf Tra. & \textbf{UAS} & \textbf{LAS} & \textbf{UPOS} & \textbf{UAS} & \textbf{LAS} & \textbf{UPOS} \\
 & \bf Size & &  & \bf F1 &  & & \bf F1  \\
\hline
$\text{STEPS}_{\text{BERTurk}}$\texttt{+TR\_BOUN} & 7,803 &$79.92_{\pm0.06}$ & $71.29_{\pm0.11}$ & $94.76_{\pm0.07}$ & 
$83.11_{\pm0.20}$ & $76.55_{\pm0.13}$ & $93.00_{\pm0.11}$ \\
$\text{STEPS}_{\text{BERTurk}}$ &&&&&&& \\
\hspace{12pt}\texttt{+TR\_BOUN}\texttt{+OTA\_BOUN}&\multirow{-2}{*}{7,917} & \bf \multirow{-2}{*}{$\mathbf{81.51_{\pm0.15}}$} & \bf \multirow{-2}{*}{$\mathbf{73.79_{\pm0.16}}$} & \bf \multirow{-2}{*}{$\mathbf{94.98_{\pm0.35}}$} & \bf \multirow{-2}{*}{$\mathbf{83.15_{\pm0.11}}$} & \bf \multirow{-2}{*}{$\mathbf{76.58_{\pm0.06}}$} & \bf \multirow{-2}{*}{$\mathbf{93.07_{\pm0.18}}$} \\
$\text{STEPS}_{\text{BERTurk}}$\texttt{+OTA\_BOUN} &114& $68.87_{\pm1.47}$ & $59.70_{\pm1.24}$ & $91.56_{\pm0.04}$ & 
$68.66_{\pm0.39}$ & $59.16_{\pm0.59}$ & $87.21_{\pm0.30}$ \\
\hline
$\text{STEPS}_{\text{mBERT}}$\texttt{+TR\_BOUN} & 7,803& $72.96_{\pm0.15}$ & $64.32_{\pm0.06}$ & $92.26_{\pm0.18}$ & 
$79.61_{\pm0.27}$ & $72.05_{\pm0.23}$ & $92.75_{\pm0.14}$ \\
$\text{STEPS}_{\text{mBERT}}$&&&&&&& \\
\hspace{12pt}\texttt{+TR\_BOUN}\texttt{+OTA\_BOUN}& \multirow{-2}{*}{7,917} & \multirow{-2}{*}{$75.86_{\pm0.15}$} & \multirow{-2}{*}{$67.87_{\pm0.11}$} & \multirow{-2}{*}{$93.12_{\pm0.16}$} & 
\multirow{-2}{*}{$79.60_{\pm0.19}$} & \multirow{-2}{*}{$72.18_{\pm0.11}$} & \multirow{-2}{*}{$92.78_{\pm0.04}$} \\
$\text{STEPS}_{\text{mBERT}}$\texttt{+OTA\_BOUN} &114 & $61.43_{\pm0.17}$ & $49.62_{\pm0.22}$ & $88.68_{\pm0.30}$ & 
$59.55_{\pm0.41}$ & $46.56_{\pm0.20}$ & $84.54_{\pm1.04}$ \\
\hline

\end{tabular}

%}

\end{table*}

When we look at the dependency parsing results, our first observation is BERTurk's superior performance over mBERT's. The best-performing model utilizing BERTurk outperforms the best mBERT-based model by almost 6\% in both UAS and LAS on historical Turkish ({\it the OTA-BOUN test set}). The gap is smaller (around 4\%) in the dependency parsing of modern Turkish sentences ({\it the TR-BOUN test set}). These results suggest that a language-specific PLM is a better option for dependency parsing of historical Turkish, even though it was pre-trained only on the modern counterpart of the language.

When we compare the parsing performance of the models trained solely on TR-BOUN (the 1st and 4th rows) with those trained exclusively on OTA-BOUN (the 3rd and 6th rows), we observe a notable advantage for the models trained on TR-BOUN. This performance difference is largely attributed to the contrasting sizes of the training sets. Models $\text{STEPS}_\text{BERTurk}$\texttt{+TR\_BOUN} and $\text{STEPS}_\text{mBERT}$\texttt{+TR\_BOUN} were trained on 7,803 modern Turkish sentences, while the training size for $\text{STEPS}_\text{BERTurk}$\texttt{+OTA\_BOUN} and $\text{STEPS}_\text{mBERT}$\texttt{+OTA\_BOUN} were limited to just 114 historical Turkish sentences.

Notably, we observe a positive effect when OTA-BOUN is combined with TR-BOUN for training. The models utilizing this combined training approach (the 2nd and 5th rows) outperform those trained only on TR-BOUN (the 1st and 4th rows) by 2.5\% and 3.5\% in LAS for BERTurk and mBERT, respectively. Although the number of historical Turkish sentences is as small as 114,  adding them to the training set made a remarkable impact on the dependency parsing of historical Turkish. These findings indicate that parsing performance for historical Turkish can be further enhanced with increased training data in historical Turkish. As anticipated, the inclusion of OTA-BOUN in the training set does not affect the parsing performance on modern Turkish sentences (TR-BOUN test set).

The positive effect of using OTA-BOUN in training is less visible in POS tagging experiments. There is almost a 1\% increase in the F1 score of the $\text{STEPS}_\text{mBERT}$
model when OTA-BOUN is added to the training set along with TR-BOUN. For the $\text{STEPS}_\text{BERTurk}$ model, OTA-BOUN has almost no effect on the performance of the model in predicting the POS tags of the OTA-BOUN test set. Here, BERTurk-based models once again outperform their corresponding mBERT-based models in this task. It is worth noting that the POS tagging models may have already approached a saturation point, as the F1 score for the best-performing model on the OTA-BOUN test set is nearly 95. In such cases, further improvements in model performance are likely to be incremental and minimal. 

All of these experimental findings indicate that leveraging a language-specific PLM trained on the modern counterpart of a historical language, followed by fine-tuning on domain-specific datasets, serves as an effective starting point for NLP tasks involving historical Turkish. Despite the limited size of labeled datasets for historical Turkish, this approach demonstrates the potential to achieve satisfactory performance on the studied tasks. However, significant challenges remain, particularly in tackling more complex tasks, such as dependency parsing, and in adapting models to out-of-domain data, as evidenced by the {\it Ruznamçe} test set.

\section{Conclusion}

This study represents a significant step forward in advancing NLP for historical Turkish. Recognizing the critical need for robust resources and tools to analyze this rich linguistic heritage, we have introduced several novel contributions: (i) the first named entity recognition dataset for historical Turkish ({\it HisTR}), enabling the identification and classification of crucial entities within historical texts; (ii) The first manually annotated dependency treebank for historical Turkish ({\it OTA-BOUN}), providing a valuable resource for syntactic analysis and model development; (iii) a clean text corpus of historical Turkish ({\it OTC}), offering a substantial foundation for various NLP tasks; (iv) trained models for dependency parsing, part-of-speech tagging, and named entity recognition tasks for historical Turkish, establishing benchmarks for future research and providing a strong starting point for further development.

By making all resources and models publicly available, we aim to foster broader research and innovation in the field of historical Turkish NLP. These contributions address the critical gap in existing resources and pave the way for more sophisticated analyses of historical documents, enabling deeper insights into the history, culture, and language of the Ottoman Empire.

As future work, we plan to expand the HisTR dataset and the OTA-BOUN treebank in terms of both size and the time periods they represent. We believe that, rather than exclusively fine-tuning PLMs on labeled historical Turkish data, first applying continual pre-training methods on PLMs using raw historical Turkish texts, followed by fine-tuning on labeled data, would significantly improve model performance for historical Turkish NLP. Therefore, we aim to enrich the OTC corpus evenly across different historical Turkish periods and pre-train PLMs on this corpus to develop better language models tailored to historical Turkish.

\paragraph{Funding Statement}
This research was supported in part by the Scientific Research Program (BAP) of Boğaziçi University under Grant 24V00SUP1.

\paragraph{Data Availability Statement}
The datasets and the trained models introduced by this work can be found in BUCOLIN Lab's HuggingFace page: \url{https://huggingface.co/BUCOLIN}.
MilliyetNER and WikiANN datasets are publicly available datasets provided at \url{https://data.tdd.ai/} and \url{https://github.com/afshinrahimi/mmner}, respectively.
TURNA NER model fine-tuned on MilliyetNER is available at \url{https://huggingface.co/boun-tabi-LMG/turna_ner_milliyet}.

\bibliographystyle{apalike}
\bibliography{sn-bibliography}% common bib file

\end{document}